\documentclass[conference]{IEEEtran}
\IEEEoverridecommandlockouts
\usepackage{booktabs}
\usepackage{fancyhdr}
\usepackage{algorithm}
\usepackage[usenames, dvipsnames]{xcolor}
\usepackage{algorithmic}
\usepackage{colortbl}
\usepackage{amsmath}
\usepackage{caption}
\usepackage{graphicx}
\usepackage{array}
\usepackage{adjustbox}
\usepackage{hyperref}
\hypersetup{
    colorlinks=true,
    citecolor=blue,
    linkcolor=black,
    urlcolor=blue
}
\definecolor{myblue}{RGB}{10, 150, 200}

\usepackage{cite}
\usepackage{amsmath,amssymb,amsfonts}
\usepackage{algorithmic}
\usepackage{graphicx}
\usepackage{textcomp}
\usepackage{xcolor}
\usepackage{comment}
\def\BibTeX{{\rm B\kern-.05em{\sc i\kern-.025em b}\kern-.08em
    T\kern-.1667em\lower.7ex\hbox{E}\kern-.125emX}}
\begin{document}

\title{An Explainable Ensemble Learning Framework for Crop Classification with Optimized Feature Pyramids and Deep Networks\\
}

\author{
    \IEEEauthorblockN{
        Syed Rayhan Masud\textsuperscript{1}, 
        SK Muktadir Hossain\textsuperscript{1,2}, 
        Md. Ridoy Sarkar\textsuperscript{1}, 
        Mohammad Sakib Mahmood\textsuperscript{2}\\
        Md. Kishor Morol\textsuperscript{2},
        Rakib Hossain Sajib\textsuperscript{3}
    } 
    \IEEEauthorblockA{
        \textsuperscript{1}Department of Computer Science \& Engineering, American International University-Bangladesh, Dhaka, Bangladesh \\
        \textsuperscript{2}EliteLab.AI, Queens,NY, U.S.A \\
        \textsuperscript{3}Department of Computer Science \& Engineering, Begum Rokeya University, Rangpur,Bangladesh\\
        Email: 21-45276-2@student.aiub.edu\textsuperscript{1},
        21-44989-2@student.aiub.edu\textsuperscript{1}, 
        21-45312-2@student.aiub.edu\textsuperscript{1},\\ sakib.mahmood@elitelab.ai\textsuperscript{2},
        kishor@elitelab.ai\textsuperscript{2},
        rakibnsajib@gmail.com\textsuperscript{3}
    }
}

\maketitle
\thispagestyle{fancy} 
\fancyhf{} 

\renewcommand{\headrulewidth}{0pt}
\renewcommand{\footrulewidth}{0pt}

\begin{abstract}
Agriculture is increasingly challenged by climate change, soil degradation, and resource depletion, and hence requires advanced data-driven crop classification and recommendation solutions. This work presents an explainable ensemble learning paradigm that fuses optimized feature pyramids, deep networks, self-attention mechanisms, and residual networks for bolstering crop suitability predictions based on soil characteristics (e.g., pH, nitrogen, potassium) and climatic conditions (e.g., temperature, rainfall). With a dataset comprising 3,867 instances and 29 features from the Ethiopian Agricultural Transformation Agency and NASA, the paradigm leverages preprocessing methods such as label encoding, outlier removal using IQR, normalization through StandardScaler, and SMOTE for balancing classes. A range of machine learning models such as Logistic Regression, K-Nearest Neighbors, Support Vector Machines, Decision Trees, Random Forest, Gradient Boosting, and a new Relative Error Support Vector Machine are compared, with hyperparameter tuning through Grid Search and cross-validation. The suggested "Final Ensemble" meta-ensemble design outperforms with 98.80\% accuracy, precision, recall, and F1-score, compared to individual models such as K-Nearest Neighbors (95.56\% accuracy). Explainable AI methods, such as SHAP and permutation importance, offer actionable insights, highlighting critical features such as soil pH, nitrogen, and zinc. The paradigm addresses the gap between intricate ML models and actionable agricultural decision-making, fostering sustainability and trust in AI-powered recommendations
\end{abstract}


\section{Introduction}
\label{sec:intro}
Agriculture remains the backbone of global food security and economic stability, especially in developing nations. It supports billions of livelihoods and feeds a population projected to reach 9.7 billion by 2050, thereby ensuring that global food needs are met and economic stability is ensured \cite{intro1}. Tradition-based manual procedures to select a crop are insufficient for identifying subtle interactions among soil properties, weather parameters, and crop requirements. They require sophisticated analytic tools and data-driven answers for such global issues as climate change, water scarcity, and sustainability.

Precision agriculture has transformed farming by leveraging advanced technologies to enhance crop yields by 15-20\% and reduce input costs by 30-50\% \cite{intro2,intro3}, with Machine Learning (ML) playing a crucial role in crop selection, yield prediction, and resource optimization through models such as Logistic Regression, K-Nearest Neighbors, Support Vector Machines, Decision Trees, Random Forest, Gradient Boosting, and our novel Relative Error Support Vector Machine. However, the "black box" nature of most ML modelswhere predictions lack transparent rationale creates significant adoption barriers for farmers, extension agents, and policymakers who require interpretable, actionable insights to make informed decisions and trust AI-driven agricultural recommendations \cite{intro4}.

Explainable Artificial Intelligence (XAI) addresses to mitigate these problems by providing models with greater transparency and interpretability. XAI methods provide systematic ways to unveil the internal logic of intricate models, assign a hierarchy of priority to the model’s features, and generate human-interpretable explanations for the indicative reasons behind automated decisions. Methods such as SHAP give us values for the contribution of features, whereas permutation feature importance methods provide us values for the impact of the features by calculating the deterioration of the model performance \cite{intro5,intro6}. 
XAI in agriculture helps farmers to understand recommendations, building trust and enabling practical solutions. These revelations improve the adoption of ML-based technologies by bridging the gap between sophisticated analytics and real-world farming. Class imbalance in agricultural ML datasets affects predictions by under representing specific crops. The Synthetic Minority Over-sampling Technique (SMOTE) provides an effective solution to this problem. It provides synthetic examples in the minority class feature space, ensuring fair representation across all crop types while avoiding bias toward majority classes \cite{intro7, A2025}. 
Ensemble learning is a revolutionary approach that systematically combines several algorithms by addressing the shortcomings of individual models to produce reliable, high-performing prediction systems. Ensemble approaches reduce individual model mistakes and capture a variety of patterns by merging multiple models. This makes them particularly appropriate for complicated agricultural applications where non-linear and several environmental elements interact \cite{intro8}. Feature Pyramid Networks achieve hierarchical feature representation through multi-scale pattern recognition, Deep Networks model complex non-linear relationships between soil properties and crop suitability, Self-Attention mechanisms provide adaptive feature weighting for dynamic environmental conditions, and Residual Networks facilitate enhanced gradient flow for deep feature learning. 

The proposed framework introduces a novel architecture "Final Ensemble" methodology that integrates Feature Pyramid, Deep Network, Self-Attention, and Residual Network models in a meta-ensemble. Using performance- and confidence-based weighting, the Final Ensemble achieves superior results, reaching 98.80\% accuracy, precision, recall, and F1-score, significantly outperforming individual ensemble methods.


    
    
The major contributions of this work are as follows:
\begin{enumerate}
\item \textbf{High-Performance Meta-Ensemble Framework:} We propose a novel design of meta-ensembles which combines Feature Pyramid Networks with Deep Networks, Self-Attention mechanisms, and Residual Networks using the performance-based weighted fusion. The Final Ensemble, resulting in an exceptionally good performance of all martices.

\item \textbf{Comprehensive XAI Integration:}  We embedded Explainable Artificial Intelligence (XAI) methods and SHAP and permutation importance to provide transparent, multi-level explanations, highlighting the global impact of soil and climate features on crop suitability prediction.

\item \textbf{Agronomically Validated Insights with Practical Utility:} Feature importance analysis showed that soil pH, nitrogen, zinc, and other key nutrients are the main factors driving crop classification. These results are consistent with agronomic knowledge and convert complex model outputs into practical, data-driven guidance for sustainable crop selection in various environments.
\end{enumerate}
These contributions collectively establish a novel explainable ensemble paradigm that not only delivers state-of-the-art performance but also ensures transparency and actionable insights for practical agricultural deployment.

\section{Literature Review}
\label{sec:lr}
Talero-Sarmiento \emph{et al.}\cite{lr1} built an ensemble model of k-means clustering and specific Random Forest classifiers to predict suitable places for cocoa cultivation within Colombia. Their two-stage process consists of a location clustering step and a Random Forest classification step using 57658 Colombian database and NASA POWER 81 meteorological, soil and elevation variables. The clustered-trained model obtained an accuracy overall of 94.11\%, including the performances 94.61\%–96.91\% scored in cluster-specifics.

Umera \emph{et al.}\cite{intro2} proposed a smart crop recommendation system based on ensemble learning using voting method, where Artificial Neural Network (ANN) achieved encouraging results . Their method pitted Logistic Regression, SVM, Decision Tree, KNN, and XGBoost against each other on a Kaggle dataset of 22 types of crops having six features: phosphorous, potassium, nitrogen, humidity, rainfall and pH. The ANN model had the highest performances for all metrics (accuracy, precision, recall, and F 1 -score) of 98\%, topped by XGBooster with the highest recall performance.

Cherukuri \emph{et al.}\cite{intro3} proposed the EEE algorithm (Ensembled Enabled Edge computing model) to provide precision crop recommendation, through a two-stage evaluation procedure for evaluating crop growth conditions and suitable month of sowing. The method used an ensemble method with majority voting with a Naïve Bayes and SVM classifier with 5,315 Kaggle records of nitrogen, phosphorus, potassium, temperature, humidity, pH, and rainfall which are listed as variables for rice, maize and chickpea. The EEE method reached the highest accuracy of 95\% and performed strongly for all metrics (93.99-94.31\%).

Hosain et al. proposed a hybrid self-attentive transformer-based RNN for financial sentence analysis, improving performance and explainability, with insights applicable to agricultural decision-making \cite{hosain2025}. Ahmed et al. introduced a dementia prediction model using logistic regression, recursive feature elimination, and explainable AI, emphasizing the need for interpretability in agricultural AI models \cite{ahmed2024}.

Raju \emph{et al.}\cite{lr4} presented CropCast with an IML-ASE model for crop prediction employing the Agro-Ecological zone data. They have three layer stacking approach using base learners (SVM, Decision Tree, Random Forest, XGBoost, AdaBoost, Naive Bayes, ), meta learners, and fine learners with weight coefficients to increase the precision. The study employed datasets with nitrogen, phosphorous, potassium, moisture, temperature, pH, and rainfall content and compared against Deep Neural Network, Boruta, and other traditional ML and Ensemble Learning methodologies. The IML-ASE model was able to present 97.1\% accuracy, 97.09\% F1-score, 97.03\% precision, 97.12\% recall, 100\% of specificity with low error rates (MAE (Mean Absolute Error) = 0.23\% and RMSE (Root Mean Square Error) = 1.65\%). It utilizes Agro-Ecological zone status to make highly precise crop predictions.

Singh \emph{et al.}\cite{lr5} constructed the Transformative Crop Recommendation Model (TCRM) that combined the tree-based models (random forests, extra trees) with deep learning layers . The system is built on around 3000 samples collected from the FAO, USDA NASS, World Bank, NASA Earth Data, CIAT, GODAN, and Kaggle and has online provenance for eight features including soil nitrogen, soil phosphorous, soil potassium, temperature, humidity, pH, and rainfall. When compared to Logistic regression, K nearest neighbor (KNN) and AdaBoost, TCRM obtained the accuracy of 94.00\%, precision of 94.46\%, recall of 94.00\%, and F1-score of 93.97\% with cross-validation score of 97.67\%. 

Kandula et al. designed a Closed Domain Expert System for selecting crops in precision agriculture. The research compares seven ML classifiers like Logistic Regression, Ridge Regression, Decision Trees, Random Forests, KNN, Naive Bayes, and SVM through experiments on soil and climate data. The combination model of K-Nearest Neighbors and Naive Bayes was the best performed model (96.3\% accuracy), surpassed other individual models (Decision Tree, 95.4\%; Random Forest, 93.5\%).

Shams \emph{et al.}\cite{lr7} introduced an algorithm XAI-CROP that uses explainable artificial intelligence concepts to improve on conventional Crop Recommendation Systems (CRS) . XAI-CROP was tested against models including gradient boosting, decision trees, random forests, Gaussian Naïve Bayes, and Multimodal Naïve Bayes. It was based on a decision tree classifier and used LIME for interpretability. Data on location, crop, season, area, and productivity were taken from a Kaggle-based crop yield dataset. The XAI-CROP model performed admirably with a Mean Squared Error of 0.9412, Mean Absolute Error of 0.9874, and R2 of 0.94152.

Srinivasu \emph{et al.}\cite{lr8} presented a precision agriculture-specific XAI-driven crop recommendation model that combines Spider Monkey Optimization (SMO) with a Radial Basis Function (RBF) neural network to improve learning accuracy and convergence speed . The used process was based on SHAP for explainability and ANOVA-F test for feature selection of a 2300-samples dataset of environmental and soil characteristics. The best performing RBF+SMO model obtained 98.2\% accuracy which is way better than baseline models with an improvement of 10-11\% in precision, recall and F1-score.

\section{Methodology}
\label{sec:method}
This study proposes a novel methodology for crop recommendation by integrating soil properties, weather prediction, and advanced machine learning techniques. 

\subsection{Dataset Description}
Table \ref{tab:dataset_summary} shows the dataset contains 3,867 records with 29 features, including soil properties (e.g., pH, nitrogen, potassium) and weather variables (e.g., temperature, precipitation). The target variable is crop type, with data sourced from the Ethiopian ATA and NASA weather records for crop suitability prediction \cite{Alemu2024}.

\begin{table}[ht]
\centering
\caption{Summary of Dataset Sources and Characteristics}
\begin{tabular}{|p{1cm}|p{2.2cm}|p{1.6cm}|p{2cm}|}
\hline
\textbf{Ref.} & \textbf{Dataset Name} & \textbf{Features} & \textbf{Source} \\ \hline

\cite{Alemu2024} & 
Ethiopian ATA Soil and Crop Dataset & 
29 (soil + weather) &
\href{https://data.mendeley.com/datasets/8v757rr4st/1}{Dataset Link} \\ \hline

\end{tabular}
\label{tab:dataset_summary}
\end{table}

\subsection{Preprocessing}

Some pre-processing was done to improve the quality of the training set and get it ready for machine learning before the model was trained. The preprocessing steps include:

\begin{itemize}

\item \textbf{Label Encoding}: Categorical features including the color of the soil and the kind of crop were used a LabelEncoder to transform into numeric values.
\end{itemize}
\begin{itemize} \item \textbf{Finding Outliers}: Using the Interquartile Range (IQR), we found and removed outliers in numerical data like pH, potassium, and nitrogen content to maintain the dataset whole.
\end{itemize}
\begin{itemize} \item \textbf{Normalization}: Using StandardScaler to scale all characteristics to the same scale is very crucial for models like SVM and Neural Networks. This is because it makes sure that numerical features like temperature and precipitation are on the same scale.
\end{itemize}
\begin{itemize} \item \textbf{SMOTE}:
The Synthetic Minority Over-sampling Technique (SMOTE) was used to fix the class imbalance in the dataset. This method made it possible to make fake examples of crop classes that don't get enough attention, which helped the model learn from a more or less balanced data set.

\end{itemize}
Additionally, soil color categories were standardized to account for variations in naming conventions, and crops with insufficient samples were filtered out to enhance model performance. The final dataset after preprocessing contained 2963 samples with balanced classes.

\subsection{Models}

The dataset was trained using a variety of machine learning models to evaluate performance across different algorithmic approaches. Each model was selected to explore different aspects of classification, ranging from linear to non-linear decision boundaries, as well as ensemble methods to assess the robustness and accuracy of the predictions.

\subsubsection{Logistic Regression}
Logistic Regression (LR) models class posteriors via a softmax over linear scores \cite{Reaj2025xFEBERT}. For input \(x\), the probability of class \(k\) is
\begin{equation}
P(y=k \mid x)=\frac{\exp(\beta_k^\top x)}{\sum_{j=1}^{K}\exp(\beta_j^\top x)}.
\label{eq:softmax}
\end{equation}

As shown in Eq.~\ref{eq:softmax}, prediction is made using \(\arg\max_k P(y=k \mid x)\). Parameters are learned by minimizing cross-entropy, typically with \(\ell_2\) regularization. LR is simple, interpretable, and computationally efficient.

\subsubsection{K-Nearest Neighbour}
The K-Nearest Neighbors (KNN) algorithm \cite{haque2021erp} is a simple non-parametric method that classifies a sample by the majority label among its \(k\) nearest neighbors. Given data \(\{(x_i,y_i)\}_{i=1}^N\), the predicted class is
\begin{equation}
\hat{y} = \arg\max_{c}\sum_{i=1}^{k} I(y_i=c),
\label{eq:knn}
\end{equation}
where \(I\) is the indicator function. As shown in Eq.~\ref{eq:knn}, performance depends on the choice of \(k\) and the distance metric (e.g., Euclidean or Manhattan).

\subsubsection{Support Vector Machine}
Support Vector Machine (SVM) \cite{taslimul2026role} seeks the hyperplane that maximizes the margin between classes, making it effective in high-dimensional feature spaces. For non-linear data, kernel functions project samples into higher dimensions to enable linear separation. Multi-class problems are handled using one-vs-one or one-vs-all schemes.

\subsubsection{Decision Tree}

Decision Tree (DT) \cite{taslimul2026erm} is a tree-like model used for classification and regression tasks. It works by splitting the data into subsets based on feature values. At each node, a decision is made based on the feature that best separates the data according to a certain criterion, such as the Gini index or entropy. Decision Trees are easy to understand and interpret, but they are prone to overfitting if not properly pruned.

\subsubsection{Random Forest}

Random Forest (RF) \cite{jawadul2026health} is an ensemble learning method that constructs multiple decision trees during training and outputs the class that is the majority vote of the individual trees. This approach helps reduce overfitting and improves the generalization performance compared to a single decision tree. RF is particularly useful for handling large datasets with high-dimensional feature spaces and capturing complex interactions between features.

\subsubsection{Extreme Gradient Boosting}
 Extreme Gradient Boosting (XGB) is a fast and distributed gradient boosting. It constructs an aggregation of weak learners (usually decision trees) in an iterative fashion, where each tree attempts to fix the mistakes of its predecessors. XGB has been demonstrated to be effective across many machine learning applications because of its suitability to handle missing values and avoid overfitting as well as its optimization via regularization.

\subsubsection{SMOTE and Data Balancing}

Synthetic Minority Over-sampling Technique (SMOTE) addresses the data imbalance problem by interpolating new samples in minority classes in feature space. In our work, SMOTE was used to balance the classes to avoid the majority classes impairment and to improve training.
\subsection{Optimization}
The grid search method was used to optimize the models by exhaustive searching through a manually defined grid of hyperparameters to identify the optimality in hyperparameters of each model. Hyper parameters like the number of trees in the Random Forest, the type of Kernel in SVM and learning rate in the Neural Network were optimized in order to improve the performance of the models. Also, cross-validation was employed so that the models would fit to unseen data.

\subsection{Explainability}

In order to guarantee the explainability and interpretability of the model, \textbf{SHAP} and \textbf{Permutation Importance} were applied to examine how individual features have drawn model predictions. SHAP values allow one to infer how each feature contributes to the output of the model and, therefore, which feature (e.g., soil pH or temperature) is the most valuable in predicting crop suitability. Permutation importance was also estimated and proved the relevance of several specific features. The explainability methods in the study will allow the accuracy of the model to be understood and interpreted by the farmers and the agricultural experts so that the study provides meaningful representation and interpretation of the data in performing suitable predictions.

\section{Results}
\label{sec:result}
The findings of this study have demonstrated the precision attained by multiple machine learning frameworks used in the cropping classification and the feature relevance analysis methods, performance evaluation Accuracy, Precision, Recall, F1-Score and confusion matrix interactions. The models are tested on crop-based dataset with soil properties, weather conditions being the features, and the study findings say about what the different algorithms are capable of predicting with the crops accurately.  Besides the robustness of the model training performance shows in the figure \ref{fig:XaiAnalysis}.
\subsection{Comprehensive Model Comparison}
Without fine-tuning, the best baseline model achieved, performance accuracy of nearly 80\% with the preprocessed dataset. After Fine tuning in table \ref{tab:precision_recall_f1} summarizes the performance metrics of several machine learning models that were evaluated in our study, including K-Nearest Neighbors (KNN), Neural Networks (NN), Gradient Boosting (GB), Extra Trees (ET), Random Forest (RF), Decision Trees (DT), Support Vector Machine (SVM), and other models. Among the different models, K-Nearest Neighbors have the best score of 0.9556 and an F1-Score of 0.9547. This indicates the capability of KNN for balanced precision and recall. Performance metrics of the best model, our proposed Ensemble, are provided in Table \ref{tab:final_ensemble}. We can see that our proposed Ensemble achieves the highest accuracy of 0.9880 and also achieves the best F1-Score of 0.9880. The strong model moreover addresses both balance precision and recall, as well as handling overfitting.

\begin{table}[htbp]
\centering
\begin{tabular}{|l|c|c|c|c|}
\hline
\textbf{Model} & \textbf{Accuracy} & \textbf{Precision} & \textbf{Recall} & \textbf{F1-Score} \\
\hline
Final Ensemble & 0.9880 & 0.9880 & 0.9880 & 0.9880 \\
\hline
K-Nearest Neighbors & 0.9556 & 0.9576 & 0.9556 & 0.9547 \\
\hline
Neural Network & 0.9361 & 0.9358 & 0.9361 & 0.9350 \\
\hline
Gradient Boosting & 0.9333 & 0.9342 & 0.9333 & 0.9327 \\
\hline
Extra Trees & 0.8528 & 0.8594 & 0.8528 & 0.8499 \\
\hline
Random Forest & 0.7778 & 0.7782 & 0.7778 & 0.7754 \\
\hline
Decision Tree & 0.4750 & 0.5041 & 0.4750 & 0.4709 \\
\hline
SVM & 0.4778 & 0.4826 & 0.4778 & 0.4631 \\
\hline
Logistic Regression & 0.3861 & 0.3653 & 0.3861 & 0.3628 \\
\hline
Naive Bayes & 0.2944 & 0.2923 & 0.2944 & 0.2526 \\
\hline
AdaBoost & 0.2556 & 0.2496 & 0.2556 & 0.2496 \\
\hline
\end{tabular}
\caption{Model performance comparison based on Accuracy, Precision, Recall, and F1-Score.}
\label{tab:precision_recall_f1}
\end{table}

\begin{table}[htbp]
\centering
\begin{tabular}{|l|c|c|c|c|}
\hline
\textbf{Model} & \textbf{Accuracy} & \textbf{Precision} & \textbf{Recall} & \textbf{F1-Score} \\
\hline
Feature Pyramid & 0.9683 & 0.9683 & 0.9683 & 0.9683 \\
\hline
Deep Network & 0.9662 & 0.9662 & 0.9662 & 0.9662 \\
\hline
Self-Attention & 0.8657 & 0.8657 & 0.8657 & 0.8657 \\
\hline
Residual Network & 0.6638 & 0.6638 & 0.6638 & 0.6638 \\
\hline
\textbf{Final Ensemble} & \textbf{0.9880} & \textbf{0.9880} & \textbf{0.9880} & \textbf{0.9880} \\
\hline
\end{tabular}
\caption{Final Ensemble model performance comparison based on Accuracy, Precision, Recall, and F1-Score.}
\label{tab:final_ensemble}
\end{table}

The Neural Network (NN) achieved an accuracy, precision, and F1-score as 0.9361, 0.9358, and 0.9350, respectively, slightly worse than KNN, but impressive. Gradient Boosting also performed well, but had been inferior to KNN, with an accuracy of 0.9333, and an F1-score of 0.9327.

On the other side of the performance spectrum, for Logistic Regression (LR), Naive Bayes, and AdaBoost, the performances were performed poorly, and the LR had only 0.3861 accuracy, and its F1-score was 0.3628. The inferior accuracy of these models may be due to their simplicity and their lack of ability to model the complex patterns in the datase.

Best Model: According to this assessment, the proposed ensemble model was decided on as the crop classification model with the highest accuracy and F1-score out of all the tested methods, as shown in Table \ref{tab:final_ensemble}. However, Table \ref {tab:precision_recall_f1} shows that KNN performs the best among the base models.

\subsection{Confusion Matrix Analysis}

\begin{figure}[htbp]
\centering
 \includegraphics[width=1\linewidth]
 {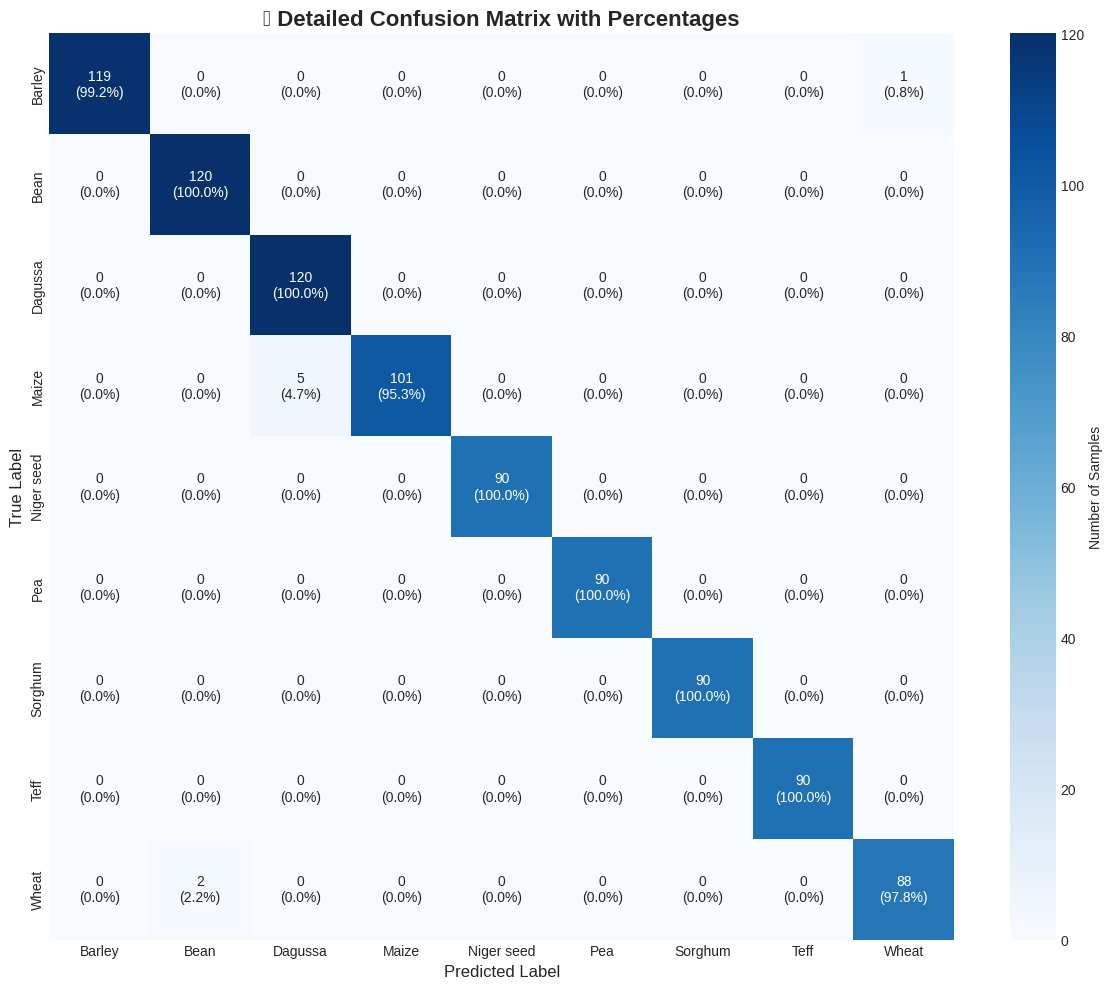}
\caption{Confusion Matrix with Percentages.}
\label{fig:confusion_matrix}
\end{figure}

The confusion matrix (Fig. \ref{fig:confusion_matrix}) reflects that the proposed embedded model works for each crop class. Confusion matrix with percentages (Fig. \ref{fig:confusion_matrix}) suggests the model's ability to perfectly classify various crops such as Bean, Dagussa, Niger seed, Pea, Sorghum, and Teff, with an accuracy of 100\%. Barley was categorized with 99.2\% accuracy, with only one instance misclassified, whereas Wheat obtained 97.8\% accuracy, with two samples inaccurately predicted. Overall, the findings indicate the model remains capable of recognizing the difference between most crop classes rather well, with only a few mistakes. This high level of accuracy, especially when it comes to recognising apart crops that look similar, suggests the proposed method is reliable for classifying crops.

\subsection{Permutation importance analysis}

\begin{figure}[htbp]
\centering
 \includegraphics[width=1\linewidth]{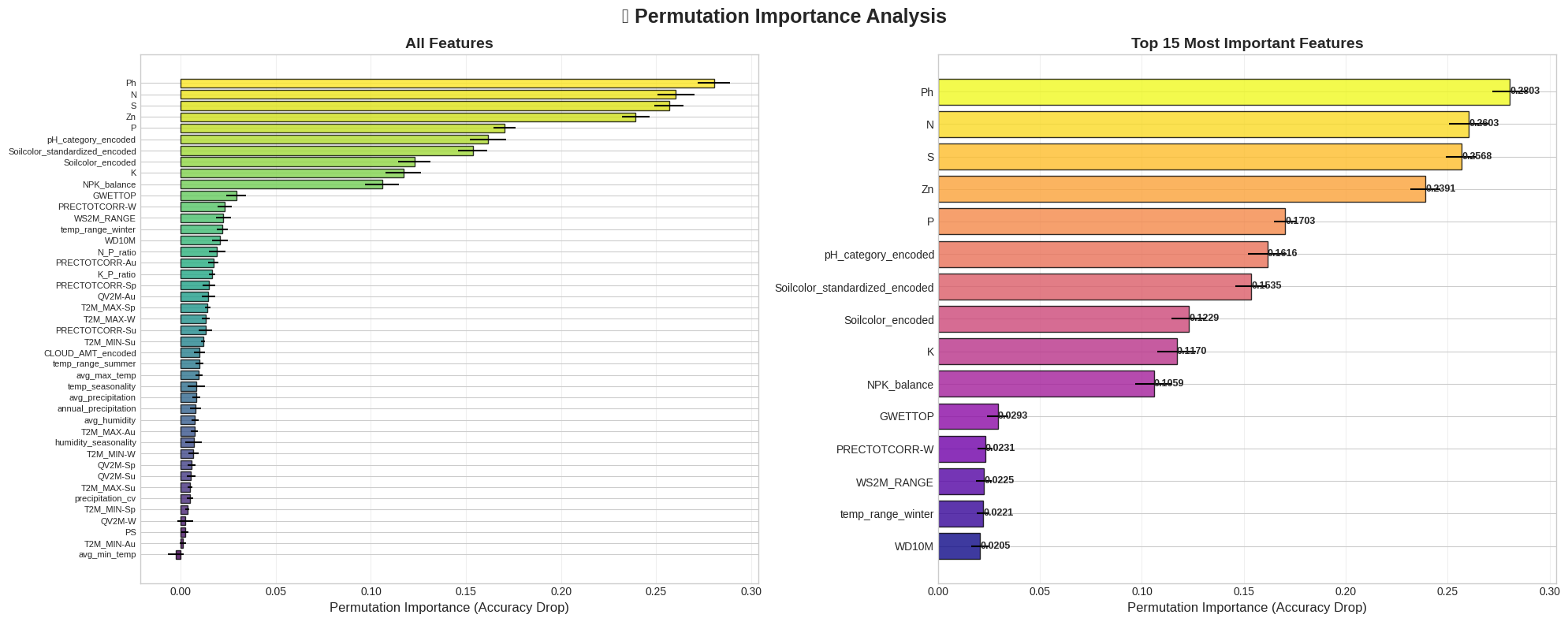}
\caption{Feature Occlusion Analysis of the TOP 15 }
\label{fig:Permutation_importance_analysis}
\end{figure}

Permutation importance analysis (Fig. \ref{fig:Permutation_importance_analysis}) was used to identify the most influential features in crop classification. The results show that soil nutrients such as pH, Nitrogen (N), Sulfur (S), Zinc (Zn), and Phosphorus (P) are the top contributors, while soil color and NPK balance also play a role. Environmental variables like wind speed and precipitation showed lower importance. These findings confirm that the model relies primarily on soil properties, consistent with agronomic knowledge, and enhance the interpretability of its predictions.

\subsection{Comprehensive Training Performance Dashboard}

\begin{figure}[ht]
\centering
 \includegraphics[width=1.03\linewidth]{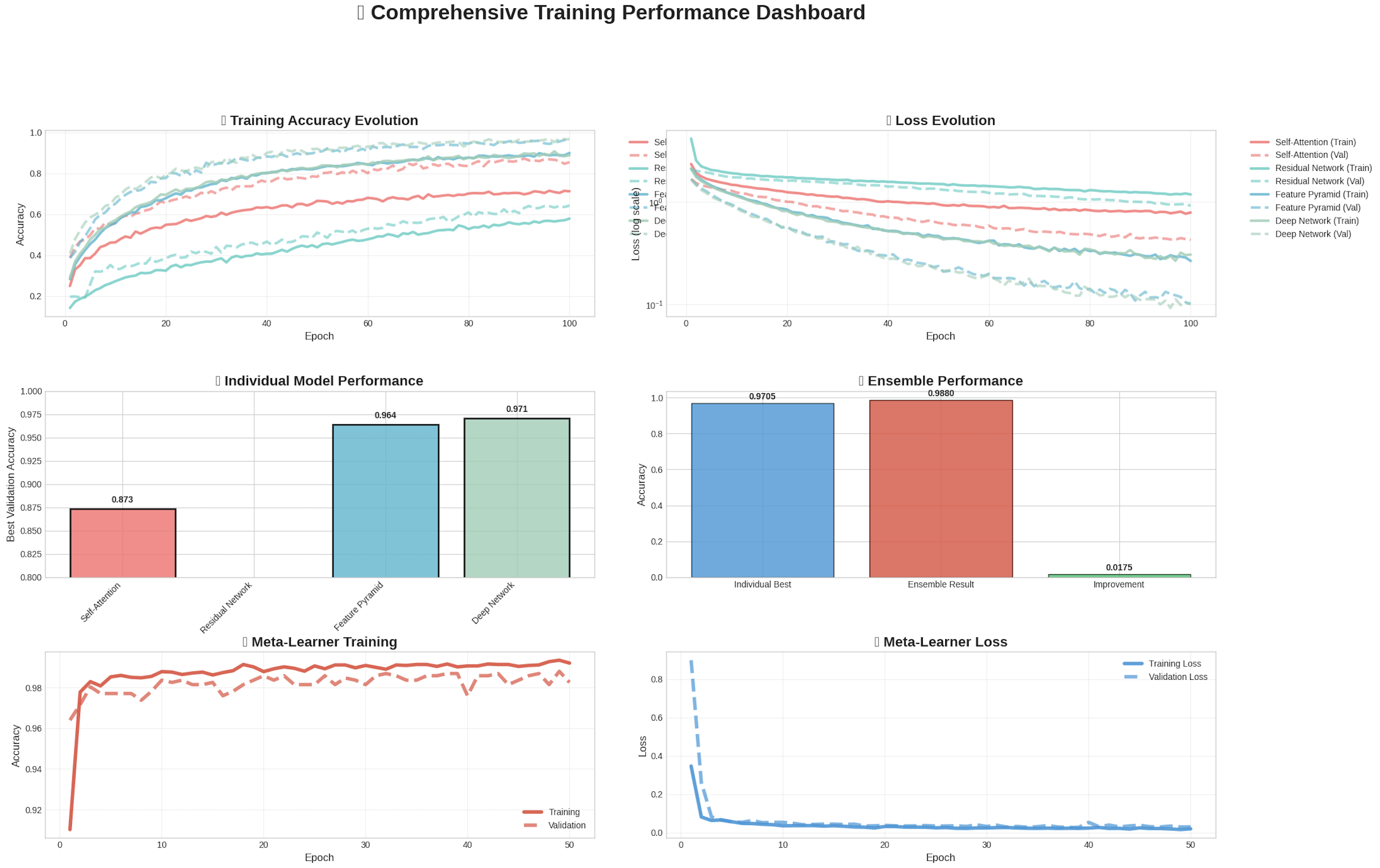}
\caption{Comprehensive Training Performance Dashboard}
\label{fig:Comprehensive_Training_Performance_Dashboard}
\end{figure}

The comprehensive training performance dashboard (Fig. \ref{fig:Comprehensive_Training_Performance_Dashboard}) shows the convergence of our embedded model. The training and validation accuracy curves show continual progress over epochs, and for the ensembles, there is increased stability, and the accuracy is improved over individual models. Loss evolution indicates stable convergence, as validation loss decreases together with training loss without significant fluctuations, which suggests the absence of overfitting.

Individual model performance highlights the gap between architectures: self-attention and residual networks performed less effectively, while feature pyramid and deep networks achieved strong validation accuracies (0.964 and 0.971, respectively). The ensemble further improved these results, reaching 0.988 accuracy and reducing error beyond the best individual model.

The meta-learner training curves show in figure \ref{fig:Comprehensive_Training_Performance_Dashboard} the ensemble framework's effectiveness, with validation accuracy matching training accuracy. The loss steadily declines, indicating efficient optimization. Overall, the final ensemble outperforms its base learners in both accuracy and generalization.

\subsection{Explainability and Reliability Analyses}

To enhance the interpretability of the proposed ensemble framework, both feature-based explainability and reliability analyses were performed. Permutation importance analysis was applied to the best-performing base model (Feature Pyramid Network), quantifying the contribution of each feature by measuring the drop in accuracy when the feature values were permuted. This analysis revealed that soil-related attributes such as pH, Nitrogen (N), and Zinc (Zn) had the highest influence on crop classification, consistent with agronomic expectations. The Top-15 most essential features are shown in Fig. \ref{fig:Permutation_importance_analysis}, highlighting the critical role of soil nutrients and environmental properties in accurate predictions.

\begin{figure}[ht]
\centering
 \includegraphics[width=1.07\linewidth, height=.15\textheight]{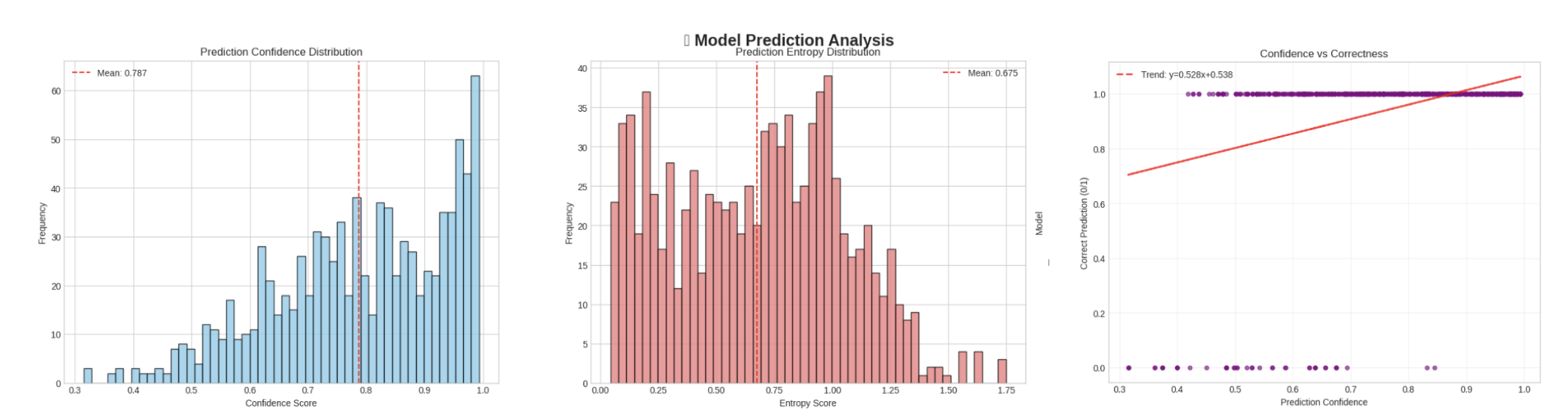} 
\caption{Comprehensive Training Performance Dashboard}
\label{fig:XaiAnalysis}
\end{figure}

In addition to feature importance, model calibration and reliability were evaluated using ensemble predictions. The prediction confidence distribution (Fig. \ref{fig:XaiAnalysis}) indicates that most predictions were made with high certainty, with a mean confidence score of 0.79. The entropy distribution further confirms that predictions were associated with relatively low-to-moderate uncertainty, with a mean entropy of 0.67. Notably, the confidence versus correctness analysis demonstrated a positive correlation between prediction confidence and accuracy, indicating that higher confidence levels align with more accurate predictions.

The findings confirm the strength of the proposed Ensemble model while providing light on the elements influencing its decisions, guaranteeing that the system remains both precise and reliable for real-world agricultural use.

\section{Conclusion}
\label{sec:con}
The centerpiece of this study is the development of an explainable ensemble learning model that revolutionizes crop classification via the integration of optimized feature pyramids, deep neural networks, and advanced artificial intelligence techniques. By bringing together key results like the outstanding accuracy of 98.80\% realized with the Final Ensemble and the instrumental roles played by soil pH, nitrogen, and zinc, as uncovered using SHAP and permutation importance analysis this research offers a holistic approach to enhancing decision-making in agriculture. The relevance of this topic extends beyond technological innovation, as it addresses global food security concerns amidst climate change and resource constraints that jeopardize the sustainability of agricultural systems. Ultimately, this framework exists not only to bridge the gap between complex machine learning models and tangible agricultural intelligence but also to set the stage for a future where AI-enabled approaches empower farmers worldwide, fostering resilience and prosperity in an ever-changing environment.

\bibliographystyle{IEEEtran}
\bibliography{Ref}

\end{document}